\documentclass[10pt,twocolumn,letterpaper]{article}

\usepackage[pagenumbers]{iccv} %

\makeatletter
\renewcommand{\maketag@@@}[1]{\hbox{\m@th\normalsize\normalfont#1}}%
\makeatother
\usepackage[ruled,noend,linesnumbered]{algorithm2e}
\usepackage{etoolbox}%
\providetoggle{beginFlag}
\settoggle{beginFlag}{true}

\SetAlgoSkip{SkipBeforeAndAfter}

\usepackage{makecell}
\usepackage{multirow}

\usepackage{bbding}

\definecolor{iccvblue}{rgb}{0.21,0.49,0.74}
\usepackage[pagebackref,breaklinks,colorlinks,allcolors=iccvblue]{hyperref}

\title{SpectralAR: Spectral Autoregressive Visual Generation}

\author{Yuanhui Huang \quad Weiliang Chen \quad Wenzhao Zheng\footnotemark[1] \\ Yueqi Duan \quad Jie Zhou \quad Jiwen Lu \\
Tsinghua University, China \\
\texttt{huangyh22@mails.tsinghua.edu.cn; wenzhao.zheng@outlook.com} 
}

\begin{document}

\twocolumn[{%
\renewcommand\twocolumn[1][]{#1}%
\vspace{-10mm}
\maketitle
\vspace{-10mm}
\begin{center}
    \centering
    \includegraphics[width=\linewidth]{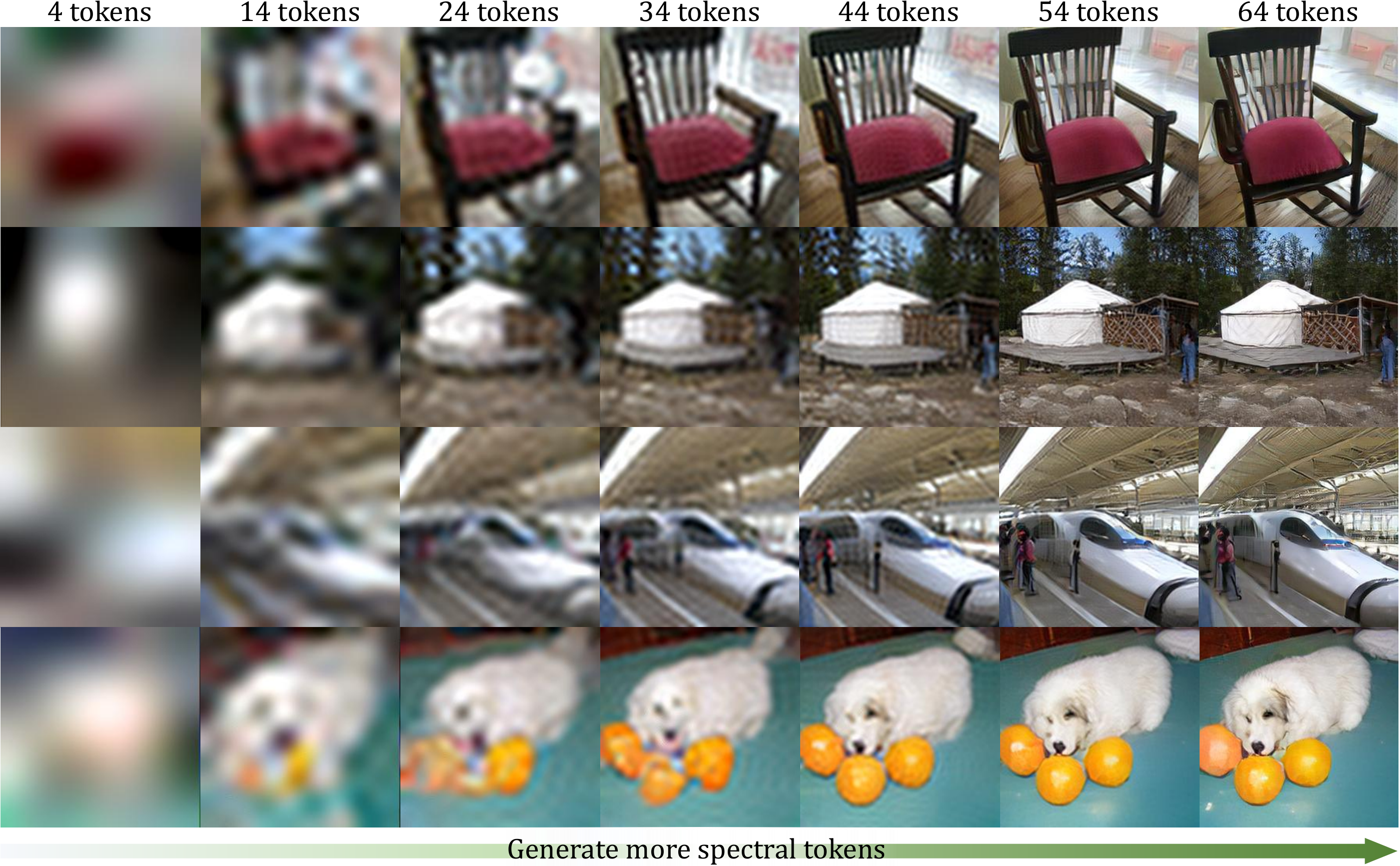}
    \vspace{-7mm}
    \captionof{figure}{
    We approach autoregressive visual generation from the spectral perspective and propose \textbf{SpectralAR} which converts images into 1D causal sequences with nested spectral tokenization and generates images in a hierarchical coarse-to-fine manner.
    In the autoregressive process, each generated token improves the quality of the image by introducing new high-frequency components.
    }
\label{teaser}
\end{center}%
}]

\renewcommand{\thefootnote}{\fnsymbol{footnote}}
\footnotetext[1]{Project leader.}
\renewcommand{\thefootnote}{\arabic{footnote}}

\begin{abstract}
Autoregressive visual generation has garnered increasing attention due to its scalability and compatibility with other modalities compared with diffusion models.
Most existing methods construct visual sequences as spatial patches for autoregressive generation.
However, image patches are inherently parallel, contradicting the causal nature of autoregressive modeling.
To address this, we propose a Spectral AutoRegressive (SpectralAR) visual generation framework, which realizes causality for visual sequences from the spectral perspective.
Specifically, we first transform an image into ordered spectral tokens with Nested Spectral Tokenization, representing lower to higher frequency components.
We then perform autoregressive generation in a coarse-to-fine manner with the sequences of spectral tokens.
By considering different levels of detail in images, our SpectralAR achieves both sequence causality and token efficiency without bells and whistles.
We conduct extensive experiments on ImageNet-1K for image reconstruction and autoregressive generation, and SpectralAR achieves 3.02 gFID with only 64 tokens and 310M parameters.
Project page: \url{https://huang-yh.github.io/spectralar/}.

\end{abstract}
    
\vspace{-5mm}
\section{Introduction}
\label{sec:intro}
Diffusion models~\cite{ho2020ddpm,song2020ddim,rombach2022ldm,peebles2023dit} have long been the best performing approach to visual generation.
Despite their exceptional generation quality, diffusion models still exhibit deficiencies in multimodal modeling and integration of perception and generation.
The advent of autoregressive visual generation methods~\cite{tian2025var,esser2021vqgan,razavi2019vqvae-2,lee2022RQ-VAE,sun2024llamavar,ma2024star,yu2024RAR} alleviates these limitations and enables better scalability with the next-token prediction paradigm.
It first utilizes a visual tokenizer to convert images into tokens and then generates samples in a sequential manner.
This advancement supports a variety of emerging applications, including scalable visual generation~\cite{wu2024janus,wang2024loong,zhang2024var-clip,han2024infinity}, mixed-modal foundation models~\cite{team2024chameleon,xie2024show-o,wang2024emu3}, and autoregressive world models~\cite{zheng2024occworld,huang2024owl}.

Despite their dominance in language modeling~\cite{vaswani2017attention,radford2018gpt}, the performance of autoregressive models in visual generation~\cite{goodfellow2014gan,esser2021vqgan} is still inferior to that of diffusion models~\cite{ma2024sit,peebles2023dit} and non-autoregressive models~\cite{chang2022maskgit,yu2025titok}.
This distinction can be attributed to the inherent difference between text and image modalities.
Text, invented for human communication, is discrete and sequential, while image data is continuous and invariant to translation~\cite{lecun1998lenet}, indicating the intrinsic equality among image pixels.
This equality makes it a critical issue for autoregressive visual generation to convert images into one-dimensional sequential tokens~\cite{yu2024RAR}.
Methods based on spatial scanning~\cite{esser2021vqgan,yu2021vim,sun2024llamagen,luo2024magvit2} attempt to discretize the image locally into tokens according to patches, and then perform autoregressive generation following a certain order based on their locations.
However, the resulting spatial sequence violates the equality among image patches, making it suboptimal for causal autoregressive modeling.
Another line of work~\cite{chang2022maskgit,li2025mar,yu2025titok} introduces bidirectional interaction in the generator as a workaround.
Nonetheless, they still tokenize images spatially, thus assuming a causal order among image patches.
In addition, the bidirectional design may deviate from the conventional autoregressive paradigm, complicating their integration into omni-modal frameworks~\cite{team2024chameleon,wang2024emu3}.
In contrast to spatial tokenization, VAR~\cite{tian2025var} explores transforming the image into multiple scales and producing a sequence by concatenating tokens from ordered scales.
Although scale-wise autoregressive generation indeed satisfies the equality of image pixels, it suffers from inferior token efficiency and parallel generation of multiple tokens from the same scale.

\begin{figure}[t!]
\centering
\includegraphics[width=\linewidth]{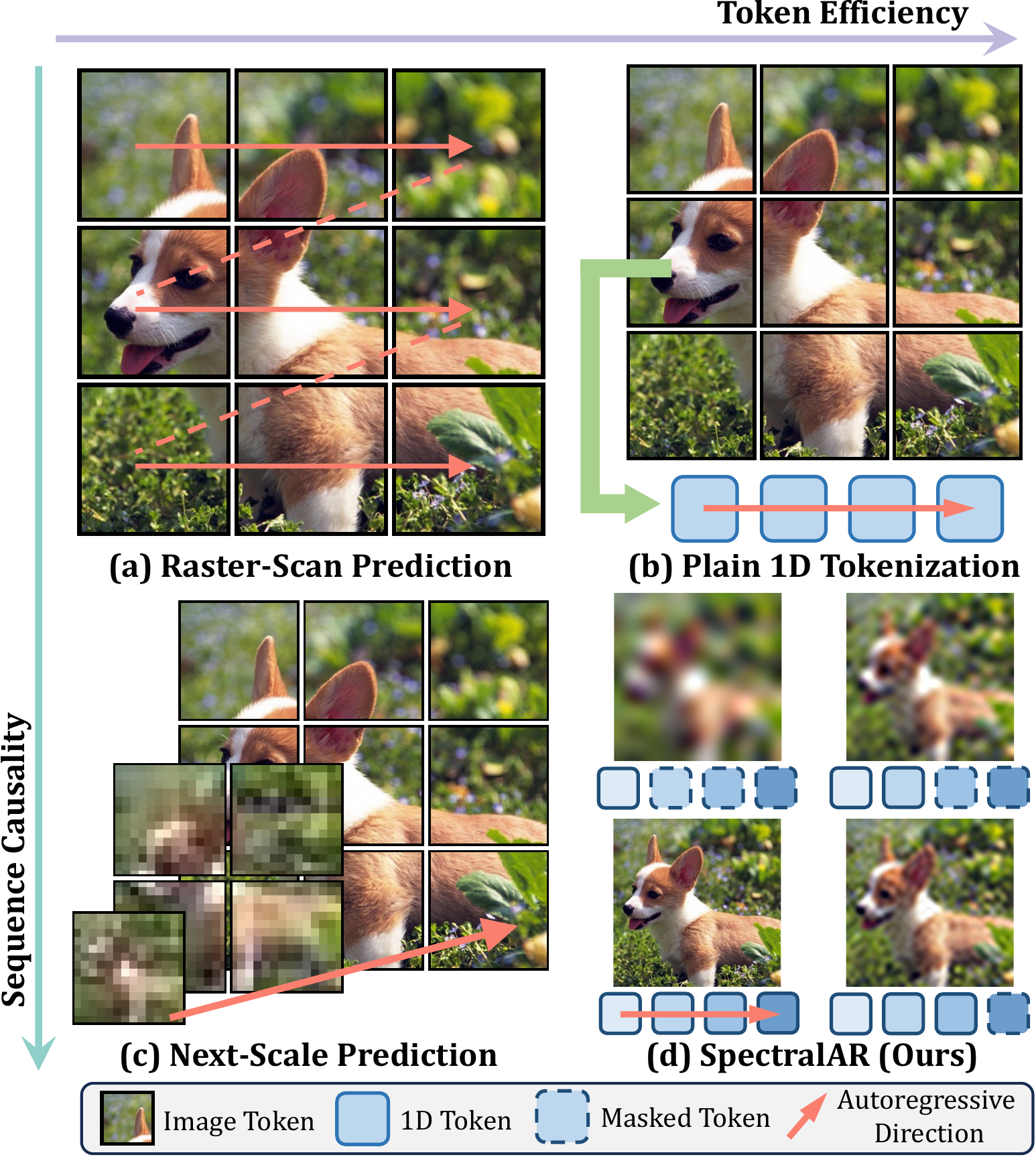}
\vspace{-7mm}
\caption{
\textbf{Comparison between autoregressive visual generation methods.}
SpectralAR achieves both token efficiency and sequence causality with nested 1D spectral tokens.
}
\label{fig:comparison}
\vspace{-7mm}
\end{figure}

In this paper, we introduce a spectral autoregressive visual generation framework to achieve causal autoregressive modeling and improve token efficiency, as shown in Figure~\ref{fig:comparison}.
Frequency is an inherent attribute of all types of signals and has become a significant perspective and methodology that complements the spatial-temporal domain~\cite{baron2003fourier}.
For visual data, the spectral density often conforms to the power-law distribution~\cite{clauset2009powerlaw}, with low-frequency components representing the overall structure of an image and high-frequency components focusing on the intricate details.
This hierarchical coarse-to-fine nature of the correspondence between spectral and spatial domains indicates a sequential order of images, motivating us to represent images as causal spectral sequences.
Specifically, we first transform images into spectral tokens with Nested Spectral Tokenization (NST), which uses varying sequence lengths to represent images across different frequency bands.
The causality of the spectral sequences originates from the coarse-to-fine progression characteristic resembling human visual perception and is strengthened with the application of a causal mask.
In addition, inspired by the image compression literature~\cite{wallace1991jpeg,wiegand2003h264}, we design a non-uniform token-frequency mapping, which allocates more tokens to represent low-frequency components and fewer for high-frequency components.
This mapping technique greatly reduces the number of tokens, while maintaining the quality of reconstructed samples.
In the autoregressive process, we begin with the token representing the DC component and progressively predict tokens corresponding to higher frequencies conditioned on previous ones.
We conduct extensive experiments on the ImageNet-1K dataset for image reconstruction and class-conditional generation.
Our SpectralAR demonstrates comparable performance with state-of-the-art methods.

\section{Related Work}
\label{sec: related work}

\textbf{Autoregressive visual generation.}
While diffusion models have achieved exceptional performance, they are fundamentally different from the conventional autoregressive framework~\cite{team2024chameleon,wang2024emu3} for cross-modal and cross-task modeling.
Therefore, a body of research aims to advance autoregressive models for visual generation.
Early efforts~\cite{van2016pixelcnn,chen2020pixels-pretraining} perform pixel-level generation in the row-major raster-scan order, followed by VQGAN~\cite{esser2021vqgan} which transfers to the latent feature space of VQVAE~\cite{van2017vqvae} for autoregressive modeling.
Subsequent work improves based on VQGAN with multiple scales~\cite{razavi2019vqvae-2}, residual quantization~\cite{lee2022RQ-VAE}, ViT architecture~\cite{yu2021vit-vqgan} or textual conditions~\cite{yu2022parti}.
However, raster-scan generation violates the equality between image pixels as discussed in Section~\ref{sec:intro}, contradicting the causality premise of autoregressive modeling.
Recently, VAR~\cite{tian2025var} proposes the scale-wise autoregressive generation that aims to predict the next-scale token map conditioned on the previous ones.
Despite achieving causality with the multi-scale design, VAR predicts multiple tokens with bidirectional attention in each step, thus deviating from the standard autoregressive framework.
In contrast, our method realizes causal autoregressive generation from the spectral perspective and still conforms to the unidirectional scheme.

\textbf{Efficient image tokenization.}
Autoregressive generation requires the conversion of images into token sequences, which is often achieved using autoencoders~\cite{hinton2006reduce-dimension,vincent2008autoencoder}.
Patch-based autoencoders~\cite{kingma2013vae,van2017vqvae,esser2021vqgan,zheng2022movq,mentzer2023fsq} tokenize images spatially, where each token corresponds to a certain patch from the original image.
Although this paradigm performs well in image reconstruction and diffusion-based generation~\cite{rombach2022ldm,peebles2023dit}, it is not suitable for autoregressive modeling due to its spatial design.
Also, its token length is proportional to the square of image resolution, which might become the bottleneck in multimodal modeling given limited context length~\cite{team2024chameleon}.
TiTok~\cite{yu2025titok} proposes a 1D tokenizer which reduces the number of tokens to 32 for $256\times 256$ images.
However, TiTok is trained with only an overall reconstruction objective, and thus the precise meaning of these 1D tokens remains unclear.
VAR~\cite{tian2025var} introduces a multi-scale tokenizer for causal autoregressive image generation.
Nonetheless, the multi-scale strategy requires an even greater number of tokens compared to patch-based tokenization methods, further diminishing token efficiency.
Our method converts images into 1D causal sequences of spectral tokens and enhances token efficiency by leveraging the long-tail distribution of spectral density in image data.

\textbf{Spectral visual analysis.}
Spectral analysis~\cite{baron2003fourier} has been a common technique in computer vision, complementing the spatial and temporal domains.
Representative applications include image enhancement and denoising~\cite{greenspan2000freq-enhance,yue2014freq-denoise}, texture analysis and feature extraction~\cite{rao2021gfnet}, compression and super-resolution~\cite{wallace1991jpeg,fritsche2019freq-superreso}, visual generation~\cite{yang2022wavegan,yu2021wavefill}, adversarial attacks and defenses~\cite{frank2020freq-defence,luo2022freq-attack,long2022freq-attack}.
For autoregressive image generation, CART~\cite{roheda2024cart} and SIT~\cite{esteves2024spectral-token} propose to transform an image into multiple causal sets of tokens with base-detail decomposition and discrete wavelet transform, respectively.
However, these methods still adhere to the multi-scale 2D tokenization paradigm similar to VAR, resulting in suboptimal token efficiency and bidirectional attention to predict multiple tokens per autoregressive step.
In contrast, SpectralAR leverages the discrete cosine transform to capture the global information of an image, and compress it into a 1D sequence with high efficiency.

\begin{figure*}[t]
\centering
\includegraphics[width=0.95\linewidth]{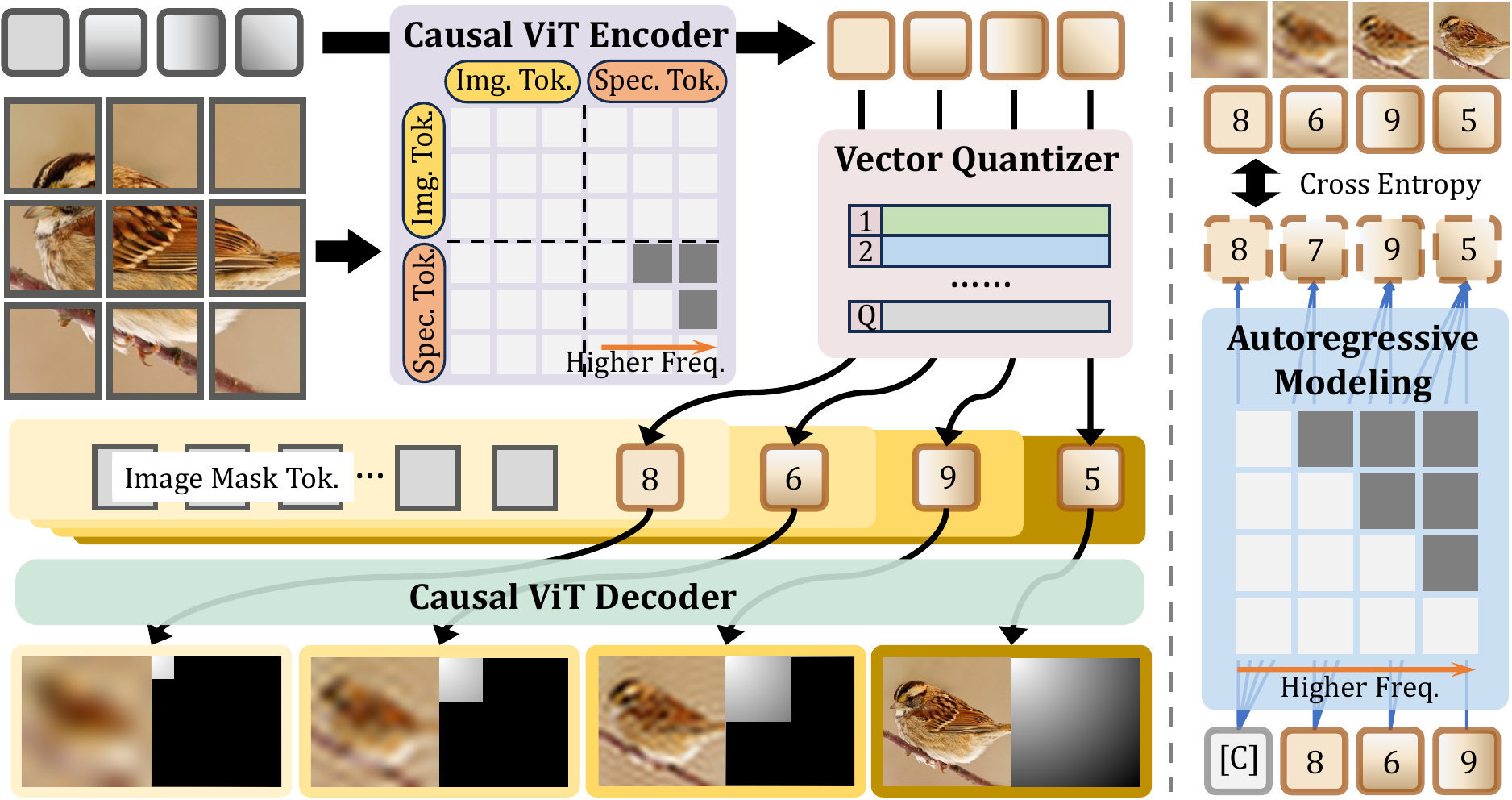}
\vspace{-3mm}
\caption{
\textbf{Overall pipeline of the proposed SpectralAR.}
Left: We convert an input image into a 1D causal sequence efficiently with nested spectral tokenization.
Each nested sequence is supervised with distinct reconstruction target in a coarse-to-fine manner, which endows each token with an explicit spectral interpretation.
We also apply the causal mask to the spectral tokens in the encoder and decoder to enhance the causality.
Right: We use the groundtruth sequences from the tokenization process to train an autoregressive generation model.
}
\label{fig:pipeline}
\vspace{-6mm}
\end{figure*}

\section{Proposed Approach}
\label{sec: approach}
In this section, we present our method of spectral autoregressive visual generation.
We first review the spectral characteristics of images and highlight the advantages of spectral autoregressive image generation (Section~\ref{subsec: 3.1}).
We then introduce our nested spectral tokenization for both causal and efficient image quantization (Section~\ref{subsec: 3.2}).
Finally, we detail the autoregressive generation process based on the sequences of spectral tokens (Section~\ref{subsec: 3.3}).

\subsection{Revisiting Images from the Spectral Domain}
\label{subsec: 3.1}
\textbf{Discrete cosine transform}.
Spectral analysis investigates how complex signals can be represented with simpler basis functions~\cite{baron2003fourier}, producing a spectral density distribution that represents the magnitude of corresponding basic components.
This spectral density distribution is an equivalent representation of the original signal and provides a distinct perspective from the spatial domain depending on the properties of the basis functions.
We employ the Discrete Cosine Transform (DCT)~\cite{ahmed2006dct} to convert images into the spectral domain.
The DCT result $\mathbf{D}$ shares the same shape with the transformed image $\mathbf{I}\in\mathbb{R}^{H\times W}$ (for simplicity, we omit the channel dimension):
\begin{equation}
    \mathbf{D} = \{F(u, v)\}_{u,v=1}^{W,H}, \quad \mathbf{I} = \{f(x, y)\}_{x,y=1}^{W,H}.
\end{equation}
Each $F(u,v)$ represents the intensity of the corresponding basis function $g_{u,v}(x,y)$ in the image $\mathbf{I}$, which writes:
\begin{small}
\begin{equation}
    g_{u,v}(x,y)=\frac{2C(u)C(v)}{\sqrt{HW}}cos\frac{(2x+1)u\pi}{2W}cos\frac{(2y+1)v\pi}{2H},
\end{equation}
\end{small}where $C(u)=1/\sqrt{2}$ if $u=0$ and $C(u)=1$ otherwise.
This family of basis functions has the following properties:
(1) Given $u$ and $v$, the basis function $g_{u,v}(x,y)$ exhibits a checkerboard-like pattern in the spatial domain, with periods along the x- and y-axis of $2W/u$ and $2H/v$, respectively.
This pattern suggests that the basis function characterizes the rate of variation of images in the spatial domain.
(2) The $F(u,v)$s also form a 2D matrix together, where the top-left corner represents the low-frequency components (small $u, v$), while the bottom-right corner corresponds to the high-frequency components (large $u, v$).

\textbf{Causality from the spectral domain.}
Since low- and high-frequency components describe overall structures and intricate details, respectively, we can decompose an image into a sequence of sub-images $\{\mathbf{I}_i'\}_{i=1}^L$ with increasing levels of detail by applying inverse-DCT on partially masked spectral density distributions $\{\mathbf{D}'_i\}_{i=1}^L$.
With more high-frequency components, the sub-images will gradually transition from blurred to sharp, as shown in Figure~\ref{fig:comparison}.
This hierarchical coarse-to-fine sequence aligns with human visual perception and artistic drawing, thus enhancing causality.

\textbf{Efficiency from the spectral domain.}
As the image compression literature~\cite{ahmed2006dct} pointed out, the DCT result $\mathbf{D}$ of images conforms to the power-law distribution with the absolute values of $F(u,v)$s on the top-left corner substantially larger than the bottom-right corner, indicating that most energy of an image is concentrated in the low-frequency components.
In addition, human visual perception is less sensitive to high-frequency signals, which have little influence on the visual quality of images.
Based on these distinctions, we can encode the high-frequency components of an image with coarser granularity to improve token efficiency, similar to the JPEG algorithm~\cite{wallace1991jpeg} which saves more than 90\% storage for images by suppressing high-frequency signals.

\subsection{Nested Spectral Tokenization}
\label{subsec: 3.2}
\textbf{Overall framework.}
In contrast to the 2D spatial tokenization that captures the local correlation between image patches, the basis function $g_{u,v}(x,y)$ of DCT encodes a global frequency pattern.
Therefore, we convert images into 1D tokens in spectral tokenization to reflect the global nature of the basis functions.
We start with the general framework of a 1D image tokenizer~\cite{yu2025titok}.
Given an image $\mathbf{I}$, we aim to encode it into $N$ discrete vectors $\mathbf{S}$, and also reconstruct the original image with $\mathbf{S}$.
We first patchify the image into $\mathbf{P}\in\mathbb{R}^{hw \times C}$, and concatenate the image features with the initial query vectors $\mathbf{S}_0$ to form $[\mathbf{P};\mathbf{S}_0]\in\mathbb{R}^{(hw+N)\times C}$, where $h$, $w$, $C$ denote the resolution of image features and the channel dimension, respectively.
We then employ the vision transformer $\mathcal{E}$ to enable feature extraction and interaction between the image features and 1D query tokens, resulting in the informative 1D representation $\hat{\mathbf{S}}$.
In the vector quantizer $\mathcal{Q}$, we match these continuous vectors with the codebook embeddings to derive the discrete representation $\mathbf{S}$, which could serve as the groudtruth for the autoregressive training.
At last, we append $\mathbf{S}$ to a set of mask tokens $\mathbf{M}\in\mathbb{R}^{hw\times C}$ and process them with the decoder network $\mathcal{D}$ similar to $\mathcal{E}$, in order to reconstruct the original image.
This overall framework could be formulated as:
\begin{equation}
    \hat{\mathbf{S}} = \mathcal{E}([\mathbf{P};\mathbf{S}_0]), \quad \mathbf{S} = \mathcal{Q}(\hat{\mathbf{S}}), \quad \hat{\mathbf{I}} = \mathcal{D}([\mathbf{M};\mathbf{S}]),
\end{equation}
where $\hat{\mathbf{I}}$ denotes the reconstructed image.
The training objective typically consists of multiple loss functions:
\begin{equation}
    \mathcal{L}_{tok} = \Vert\hat{\mathbf{I}} - \mathbf{I}\Vert_2^2 + \Vert\hat{\mathbf{S}} - \mathbf{S}\Vert_2^2 + \mathcal{L}_P(\hat{\mathbf{I}},\mathbf{I}) + \mathcal{L}_A(\hat{\mathbf{I}},\mathbf{I}),
\end{equation}
where $\mathcal{L}_P$ and $\mathcal{L}_A$ denote the perceptual loss~\cite{zhang2018lpips,johnson2016lpips} and the adversarial loss~\cite{goodfellow2014gan,esser2021vqgan}, respectively.

\textbf{Nested spectral decoding.}
Different from the plain 1D tokenization in Figure~\ref{fig:comparison}, we aim to represent an image as its spectral decompositions $\{\mathbf{I}'_i\}_{i=1}^L$ for sequence causality, which requires establishing a mapping from 1D tokens to these sub-images.
One naive way to achieve this would be dividing $\mathbf{S}$ into disjoint subsets and assigning them to model different sub-images independently.
Similar to the multi-scale tokenization~\cite{tian2025var}, this strategy would inevitably involve bidirectional interaction and diminish token efficiency because sub-images with finer detail would require increasingly more tokens to represent.
In contrast, we propose a nested mapping scheme for efficient tokenization, as shown in the bottom of Figure~\ref{fig:pipeline}.
We first construct a sequence of sub-images with increasing detail by progressively preserving larger regions in the spectral density $\mathbf{D}$:
\begin{equation}
    \mathbf{I}'_i={\rm DCT}^{-1}(\mathbf{D}'_i), \quad \mathbf{D}'_i=\mathbf{D} \circ \mathbf{1}_{\omega_i}, \quad \omega_{i-1} < \omega_i,
\label{eq:sub-image construction}
\end{equation}
where ${\rm DCT}^{-1}$, $\circ$, $\mathbf{1}_{\omega_i}$ denote the inverse DCT operation, element-wise multiplication and a $H\times W$ matrix with the top-left corner of size $\omega_i\times\omega_i$ filled by ones and the remaining parts being zeros.
Therefore, the sub-image $\mathbf{I}'_i$ contains all frequency components present in the sub-image $\mathbf{I}'_{i-1}$.
Based on this inclusion property, we can reuse the tokens representing the previous sub-image $\mathbf{I}'_{i-1}$ to represent the next sub-image $\mathbf{I}'_i$.
To avoid bidirectional attention, we make $L$ equal to $N$ so that each token $\mathbf{s}_i$ in the sequence $\mathbf{S}$ corresponds to a unique sub-image $\mathbf{I}'_i$:
\begin{equation}
    \hat{\mathbf{I}}'_i = \mathcal{D}([\mathbf{M};\mathbf{s}_1,\mathbf{s}_2,...,\mathbf{s}_i]),
\label{eq:nested decoding}
\end{equation}
where $\hat{\mathbf{I}}'_i$ is the reconstruction of sub-image $\mathbf{I}'_i$ given the nested 1D sequence $\{\mathbf{s}_1,\mathbf{s}_2,...,\mathbf{s}_i\}$.
Nested spectral decoding compresses an image into a causal 1D sequence where each token $\mathbf{s}_i$ corresponds to a disjoint set of frequencies and achieves token efficiency by reusing previous tokens.

\begin{algorithm}[t]
\caption{\small{~Nested Spectral Tokenization Training}} \label{alg:enc}
\small{
\textbf{Inputs: } raw image $\mathbf{I}$, initial spectral tokens $\mathbf{S}_0\in\mathbb{R}^{N\times C}$\;
\textbf{Hyperparameters: } spectral levels $\{\omega_i|i=1,...,N\}$\;
$\mathbf{P}=\text{patchify}(\mathbf{I})$, $\hat{\mathbf{S}} = \mathcal{E}([\mathbf{P};\mathbf{S}_0])$, $\mathbf{S}=\mathcal{Q}(\hat{\mathbf{S}})$\;
$idx = \text{random\_choice}(N)$\;
$\mathbf{S}'=\mathbf{S}[:idx]$, $\hat{\mathbf{I}}'=\mathcal{D}([\mathbf{M};\mathbf{S}'])$\;
$\mathbf{D} = \text{DCT}(\mathbf{I})$, $\mathbf{D}'=\mathbf{D}\circ\mathbf{1}_{\omega_i}$, $\mathbf{I}'=\text{DCT}^{-1}(\mathbf{D'})$\;
$loss=\mathcal{L}_{tok}(\hat{\mathbf{I}}',\mathbf{I}')$\;
\textbf{Return: } $loss$ for optimization\;
}
\end{algorithm}

\textbf{Non-uniform token-frequency mapping.}
To further enhance token efficiency, we introduce the non-uniform token-frequency mapping technique.
Since high-frequency components have low magnitude and minimal impact on the visual quality of images, we can encode them with coarser granularity compared to the low-frequency counterparts.
We achieve this by reducing the interval between $\omega_{i-1}$ and $\omega_i$ when $i$ is small, and increase it otherwise:
\begin{equation}
    \omega_i - \omega_{i-1} \le \omega_{i+1} - \omega_i,
\end{equation}
which demonstrates the general case.
This non-uniform mapping allocates later tokens to broader frequency ranges, enabling precise modeling of crucial low-frequency components while efficiently representing high-frequency details.

\textbf{Spectral causal mask.}
Although the sequence $\mathbf{S}$ supervised with (\ref{eq:sub-image construction})(\ref{eq:nested decoding}) already exhibits a certain degree of causality, the encoding and decoding processes, i.e. $\mathcal{E}$ and $\mathcal{D}$, are still bidirectional, which can lead to information leakage from high-frequency to low-frequency components.
Therefore, we propose applying causal masks to the spectral tokens $\mathbf{S}$ in both the encoder $\mathcal{E}$ and the decoder $\mathcal{D}$, as shown in Figure~\ref{fig:pipeline}.
This spectral causal mask restricts each token $\mathbf{s}$ to only attend to tokens that represent lower frequencies, enhancing causality from the architectural perspective.
We outline the training procedure in Algorithm~\ref{alg:enc}.

\begin{table}[t]
    \centering
    \caption{
    \textbf{Correlation between tokens of different autoregressive paradigms.}
    We use linear correlation as a proxy metric for the causality of sequences.
    The spectral sequence demonstrates better causality compared with other methods.
    }
    \vspace{-3mm}
    \setlength{\tabcolsep}{0.03\linewidth}
    \resizebox{1\linewidth}{!}{
    \begin{tabular}{l|ccc}
        \toprule
        Correlation Type & Raster-scan & Scale-wise & Spectral \\
        \midrule
        $R^2_{avg}(\mathbf{t}_2; \mathbf{t}_1)$ & 0.471 & 0.889 & \textbf{0.916} \\
        $R^2_{avg}(\mathbf{t}_3; \mathbf{t}_1, \mathbf{t}_2)$ & 0.366 & 0.953 & \textbf{0.977} \\
        $R^2_{avg}(\mathbf{t}_4; \mathbf{t}_1, \mathbf{t}_2, \mathbf{t}_3)$ & 0.525 & 0.943 & \textbf{0.994} \\
        \midrule
        Average & 0.454 & 0.928 & \textbf{0.962} \\
        \bottomrule
    \end{tabular}}
    \vspace{-3mm}
    \label{tab:poc}
\end{table}

\begin{figure}[t]
\centering
\includegraphics[width=\linewidth]{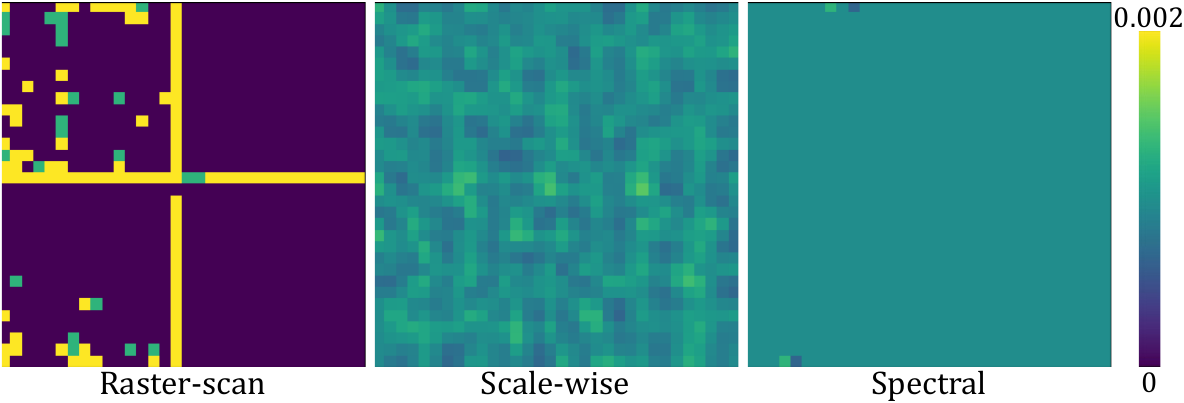}
\vspace{-7mm}
\caption{
\textbf{Frequency of each element having the highest correlation with another one.}
It represents the source of reference information for predicting the next token.
The raster-scan method exhibits excessive dependency on boundaries of image patches.
}
\label{fig:poc}
\vspace{-6mm}
\end{figure}

\begin{table*}[t]
\renewcommand\arraystretch{1.0}
\centering
\setlength{\tabcolsep}{0.013\linewidth}
\caption{
\textbf{Comparison between generative models on class-conditional ImageNet 256$\times$256 benchmark.}
``$\downarrow$'' or ``$\uparrow$'' indicate lower or higher values are better.
``\#Token'': the number of tokens used in transformer architectures.
``\#Step'': the number of model runs needed to generate an image.
We compute our wall-clock inference time relative to VAR~\cite{tian2025var} and scale the Time for other methods accordingly.
$\dag$: trained on larger datasets including OpenImages~\cite{kuznetsova2020openimages}.
$\ddagger$: implemented with the official tokenizer weight~\cite{yu2025titok} and the scripts from VAR.
}
\label{tab:main}
\vspace{-3mm}
\resizebox{0.95\linewidth}{!}{
\begin{tabular}{c|l|c|cccc|ccc|c}
\toprule
Type & Model & rFID$\downarrow$ & gFID$\downarrow$ & IS$\uparrow$ & Pre$\uparrow$ & Rec$\uparrow$ & \#Token & \#Para & \#Step & Time \\
\midrule
GAN   & BigGAN~\cite{brock2018biggan} & 75.24 & 6.95  & 224.5 & \textbf{0.89} & 0.38 & \multirow{3}{*}{N.A.} & 112M & 1    & $-$    \\
GAN   & GigaGAN~\cite{kang2023gigagan} & $-$ & 3.45  & 225.5 & 0.84 & {0.61} &  & 569M & 1    & $-$ \\
GAN   & StyleGan-XL~\cite{sauer2022styleganxl} & 7.06  & 2.30  & 265.1 & 0.78 & 0.53 &  & 166M & 1    & 0.4 \\ %
\midrule
Diff. & ADM~\cite{dhariwal2021guided-diffusion} & 125.78 & 10.94 & 101.0 & 0.69 & \textbf{0.63} & N.A.  & 554M & 250  & 235 \\ %
Diff. & CDM~\cite{ho2022cdm} & $-$ & 4.88  & 158.7 & $-$  & $-$ & N.A.   & $-$  & 8100 & $-$    \\
Diff. & LDM-4-G~\cite{rombach2022ldm} & 0.27$^\dag$ & 3.60  & 247.7 & $-$  & $-$ & N.A.   & 400M & 250  & $-$    \\
Diff. & DiT-L/2~\cite{peebles2023dit} & 0.62$^\dag$ & 5.02  & 167.2 & 0.75 & 0.57 & 256  & 458M & 250  & 43     \\
Diff. & DiT-XL/2~\cite{peebles2023dit} & 0.62$^\dag$ & 2.27  & 278.2 & 0.83 & 0.57 & 256  & 675M & 250  & 63     \\
Diff. & L-DiT-3B~\cite{zhang2023llamaadapter} & $-$ & 2.10  & 304.4 & 0.82 & 0.60 & 256  & 3.0B & 250  & $>$63     \\
Diff. & L-DiT-7B~\cite{zhang2023llamaadapter} & $-$ & 2.28  & 316.2 & 0.83 & 0.58 & 256  & 7.0B & 250  & $>$63     \\
\midrule
Mask. & MaskGIT~\cite{chang2022maskgit} & 2.28  & 6.18  & 182.1 & 0.80 & 0.51 & 256 & 227M & 8    & 0.7 \\ %
Mask. & RCG (cond.)~\cite{li2023rcg} & $-$ & 3.49  & 215.5 & $-$  & $-$ &  $-$ & 502M & 20  & 2.7 \\ %
Mask. & TiTok-B64~\cite{yu2025titok} & 1.70 & 2.48  & 214.7 & $-$  & $-$ & 64  & 177M & 8 & 0.4 \\
\midrule
2D Scan   & VQVAE-2~\cite{razavi2019vqvae-2} & $-$  & 31.11 & $\sim$45 & 0.36 & 0.57 & N.A. & 13.5B    & 5120    & $-$  \\
2D Scan    & VQGAN~\cite{esser2021vqgan} &  7.94 & 15.78 & 74.3   & $-$  & $-$ & N.A.  & 1.4B & 256  & 34     \\
2D Scan    & ViTVQ~\cite{yu2021vit-vqgan} & 1.28 & 4.17  & 175.1  & $-$  & $-$ & 1024  & 1.7B & 1024  & $>$34     \\
2D Scan    & RQTran.~\cite{lee2022RQ-VAE} &  3.20 & 7.55  & 134.0  & $-$  & $-$ & 64,\ 4  & 3.8B & 68  & 29    \\
\midrule
VAR   & VAR-$d16$~\cite{tian2025var} & \multirow{4}{*}{0.90$^\dag$}      & 3.30  & 274.4 & 0.84 & 0.51 & \multirow{4}{*}{680} &  310M & 10   & 0.6      \\
VAR   & VAR-$d20$~\cite{tian2025var} &       & 2.57  & 302.6 & 0.83 & 0.56 &  &  600M & 10   & 0.7      \\
VAR   & VAR-$d24$~\cite{tian2025var} &       & 2.09  & 312.9 & 0.82 & 0.59 &  &  1.0B & 10   & 0.8      \\
VAR   & VAR-$d30$~\cite{tian2025var} &       & \textbf{1.92}  & \textbf{323.1} & 0.82 & 0.59 &  &  2.0B & 10   & 1.4      \\
\midrule
1D AR & TiTok-B64-$d16$$^\ddagger$~\cite{yu2025titok} & 1.70 & 6.30 & 190.1 & 0.85 & 0.47 & 64 & 310M & 64 & 1 \\
1D AR & \textbf{SpectralAR}-$d16$ & 4.03 & 3.02 & 282.2 & 0.81 & 0.55 & 64 & 310M & 64 & 1 \\
1D AR & \textbf{SpectralAR}-$d20$ & 4.03 & 2.49 & 305.4 & - & - & 64 & 600M & 64 & 1.2 \\
1D AR & \textbf{SpectralAR}-$d24$ & 4.03 & 2.13 & 307.7 & - & - & 64 & 1.0B & 64 & 1.4 \\
1D AR & \textbf{SpectralAR}-$d16$-p4 & 4.03 & 3.13 & 276.1 & - & - & 16$\times$4 & 310M & 16 & 0.4 \\
\bottomrule
\end{tabular}}
\vspace{-5mm}
\end{table*}

\subsection{Spectral Autoregressive Generation}
\label{subsec: 3.3}
Autoregressive modeling has gained prominence in computer vision for its scalability, generalization, and effectiveness across multimodal tasks~\cite{team2024chameleon,wang2024emu3,kondratyuk2023videopoet}.
While conventional autoregressive generation follows a spatial raster-scan order~\cite{esser2021vqgan,van2016pixelcnn}, we propose a hierarchical coarse-to-fine approach in the spectral domain to enhance causality, as shown in the right side of Figure~\ref{fig:pipeline}.
We start with the general framework of autoregressive modeling:
\begin{equation}
    \mathbf{p}_{i+1} = \mathcal{M}(\mathbf{t}_1, \mathbf{t}_2, ..., \mathbf{t}_i),
    \label{eq:ar}
\end{equation}
where $\mathbf{T}=\{\mathbf{t}_i\}_{i=1}^{N}$ is a sequence of quantized tokens, and $\mathbf{p}_{i+1}$, $\mathcal{M}$ denote the probability logits for the (i+1)th token and the autoregressive model, respectively.
The autoregressive process (\ref{eq:ar}) assumes that the generation of token $\mathbf{t}_{i+1}$ depends solely on the previous ones.
The spatial autoregressive paradigm violates this premise because of the equality of image pixels (as discussed in Section~\ref{sec:intro}).
In contrast, we take the spectral tokens $\mathbf{S}$ from the nested spectral tokenziation as the autoregressive targets $\mathbf{T}$.
Since the spectral tokens are trained in a nested manner to reconstruct sub-images of increasing levels of detail as in (\ref{eq:sub-image construction})(\ref{eq:nested decoding}), each spectral token $\mathbf{s}_i$ is expected to enhance the quality of the sub-image $\mathbf{I}'_{i-1}$ represented by previous tokens from the spectral domain.
This progressive refinement process aligns with human visual perception and artistic painting, both of which start with the overall structure and gradually focus on details.
This similarity qualitatively validates the rationale for performing causal autoregressive generation in the spectral domain.
We further provide some quantitative analysis through a toy experiment in Section~\ref{subsec: poc}.

\textbf{Potential applications.}
The frequencies represented by the token $\mathbf{s}_i$ become higher as $i$ increases, while its influence on the image quality diminishes accordingly (check Section~\ref{subsec: 3.1} for details).
Therefore, we can control the visual quality of sampled images by adjusting the length of generated sequences, similar to the image compression algorithms~\cite{wallace1991jpeg}.
In addition, we can achieve super-resolution by dividing images into disjoint parts smaller than $H\times W$, and conducting individual spectral autoregressive generation on each part.
We provide further results in Section~\ref{subsec:app}.
\vspace{-5mm}

\section{Experiments}
\label{sec: exp}

\begin{figure*}[t]
\centering
\includegraphics[width=\linewidth]{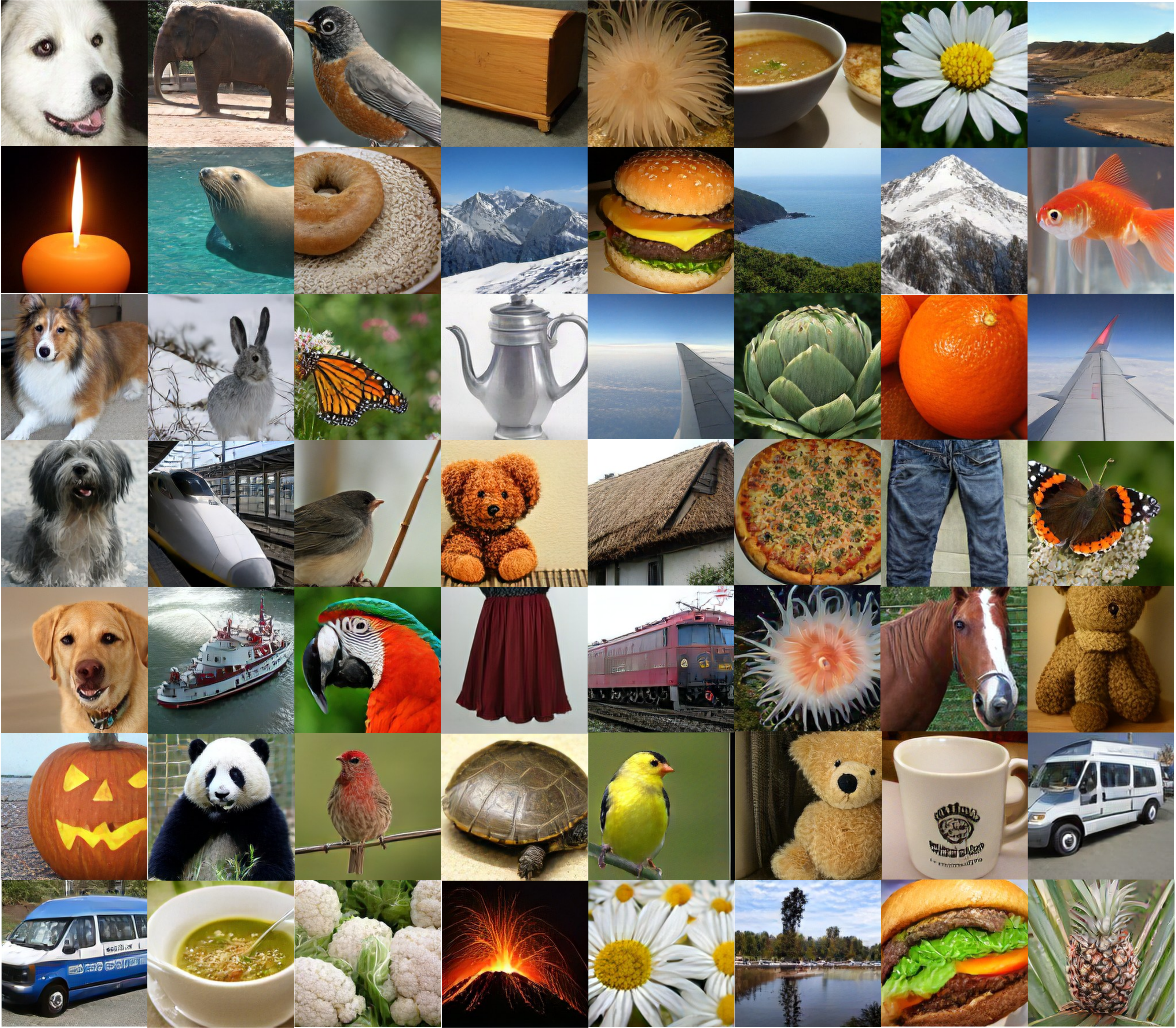}
\vspace{-7mm}
\caption{
\textbf{Visualizations of generated 256$\times$256 samples on ImageNet-1K.}
These samples cover a wide range of categories and styles, demonstrating the ability of SpectralAR to generate both diverse and high-quality images.
}
\label{fig:vis1}
\vspace{-6mm}
\end{figure*}

\subsection{Dataset and Implementation Details}
We train and evaluate our SpectralAR on the large scale ImageNet-1K~\cite{deng2009imagenet} generation benchmark, which contains 1,281,167, 50,000 and 100,000 images for training, validation and testing, respectively.
We train our tokenizer and generator only on the training split.
We evaluate the reconstruction performance of the tokenizer on the validation set with reconstruction Fréchet inception distance~\cite{heusel2017fid} (rFID), and the generation results with generation FID (gFID) using pre-computed statistics and scripts from ADM~\cite{dhariwal2021guided-diffusion}.

For tokenizer training, we follow the exactly same settings of TiTok~\cite{yu2025titok} for a fair comparison.
We also employ the two-stage training strategy with proxy codes~\cite{yu2025titok,chang2022maskgit}.
We use the ViT-B~\cite{dosovitskiy2020vit} as the encoder and decoder, and set the number of spectral tokens as $N=64$ in our main experiments and the sequence $\omega$s for $256\times 256$ images as:
\begin{equation}
    \omega_i=\left\{
        \begin{array}{ll}
        i     , & if\ \ i\in (0, 32], \\
        2i-32  , & if\ \ i\in (32, 48], \\
        12i-512, & if\ \ i\in (48, 64].
        \end{array}
    \right.
\end{equation}
For generator training, we adopt the same architecture and training recipe as VAR~\cite{tian2025var}, which leverages a GPT-2-like transformer architecture~\cite{radford2018gpt} for autoregressive modeling.

\subsection{Proof of Concept}
\label{subsec: poc}
We conduct a proof-of-concept experiment to validate the causality of images in the spectral domain.
We use the validation set of the CIFAR-100 dataset~\cite{krizhevsky2009cifar} as a proxy for the distribution of natural images, which contains 10,000 32$\times$32 images from 100 semantic classes.
We aim to measure the linear correlation ($R^2$)~\cite{ozer1985rsquare} between one token $\mathbf{t}_i$ and its previous ones $\{\mathbf{t}_j\}_{j=1}^{i-1}$ as a metric for the causality of the sequence $\mathbf{T}$.
We consider three typical autoregressive paradigms, including raster-scan~\cite{esser2021vqgan}, scale-wise~\cite{tian2025var} and spectral methods which construct sequence $\mathbf{T}$ in different ways.
Specifically, we divide each image spatially into four patches of size 16$\times$16 and rearrange them in raster-scan order for the first variant.
For the scale-wise method, we downsample each image into sizes of 8$\times$8, 16$\times$16, 24$\times$24 and 32$\times$32, respectively.
We set the frequency thresholds to 8, 16, 24, 32 and generate the sub-images following (\ref{eq:sub-image construction}) for the spectral sequence.
Since each token $\mathbf{t}$ is a high dimensional tensor, we calculate $R^2$ between every element of $\mathbf{t}_i$ and every element of $\{\mathbf{t}_j\}_{j=1}^{i-1}$.
For each element of $\mathbf{t}_i$, we then find the element in $\{\mathbf{t}_j\}_{j=1}^{i-1}$ with the highest correlation, and take an average as the final measurement:
\begin{equation}\vspace{-2mm}
    R^2_{avg} = \mathop{\rm AVG}_m(\max_n\ R^2_{m,n}),
    \label{eq:rsquare}
\end{equation}
where $m$, $n$ denote element number of $\mathbf{t}_i$ and $\{\mathbf{t}_j\}_{j=1}^{i-1}$, respectively, and $R^2_{m,n}$ is the correlation between the $m$th and $n$th elements.
We report the quantitative results in Table~\ref{tab:poc}, where the scale-wise and spectral sequences demonstrate considerably higher correlation than the raster-scan counterpart.
This could be attributed to the coarse-to-fine nature of the former two paradigms, while the raster scan method lacks adequate reference information to predict the next image patch, as shown in Figure~\ref{fig:poc}.

\begin{table}[t]
\renewcommand\arraystretch{1.0}
\centering
\setlength{\tabcolsep}{0.030\linewidth}
\caption{
\textbf{Applications of SepctralAR.}
Super-reso., Trunc. represent super-resolution and truncated, respectively.
}
\label{tab:app}
\vspace{-3mm}
\resizebox{\linewidth}{!}
{
\begin{tabular}{c|l|cc}%
\toprule
App. Type & Model & FID$\downarrow$ & IS$\uparrow$ \\
\midrule
\multirow{3}{*}{Super-reso.} & Upsample & 3.09 & \textbf{286.6} \\
& SpectralAR-Stride & \textbf{2.93} & 276.4 \\
& SpectralAR-Patch   & 14.76  & 170.0 \\
\midrule
\multirow{3}{*}{Trunc.} & SpectralAR-Trunc.5 & 3.34  & 271.5  \\
& SpectralAR-Trunc.10 & 6.65 & 211.4  \\
& SpectralAR-Trunc.15 & 27.69  & 91.69 \\
& SpectralAR-Trunc.0  & \textbf{3.02} & \textbf{282.2} \\
\bottomrule
\end{tabular}
}
\vspace{-5.5mm}
\end{table}

\subsection{Main Results}
We report the performance of SpectralAR on the class-conditional ImageNet-1K~\cite{deng2009imagenet} 256$\times$256 generation benchmark in Table~\ref{tab:main}.
We also implement an autoregressive version of TiTok~\cite{yu2025titok} by using the official tokenizer weight and scripts from VAR~\cite{tian2025var} for a fair comparison.
The reconstruction performance of SpectralAR (4.03 rFID) is inferior compared with TiTok because SpectralAR requires to reconstruct the sub-images corresponding to different frequencies with different lengths of tokens, which is much more difficult compared with the overall target in TiTok. 
Despite the lower reconstruction score, SpectralAR still outperforms TiTok-B64-$d16$ and VAR-$d16$ in autoregressive generation with a clear margin due to better sequence causality which eases autoregressive learning.
In addition, SpectralAR uses only 64 tokens in both reconstruction and generation, demonstrating superior token efficiency compared to VAR~\cite{tian2025var} and 2D scan-based methods.
We also visualize the generated samples in Figure~\ref{fig:vis1}, which demonstrate the diversity and quality of the generation process of SpectralAR.
In addition, Figure~\ref{teaser} highlights the hierarchical coarse-to-fine refinement of the images, while SpectralAR generates more tokens in an autoregressive way.

\begin{table}[t]
\renewcommand\arraystretch{1.0}
\centering
\setlength{\tabcolsep}{0.027\linewidth}
\caption{
\textbf{Ablation on design choices.}
Spectral Supervision means using sub-image supervision across diffirent frequency bands.
Causal Mask and non-uniform mapping refers to the spectral causal mask and the frequency-token mapping, respectively.
}
\label{tab:ablate}
\vspace{-3mm}
\resizebox{\linewidth}{!}
{
\begin{tabular}{ccc|cc}%
\toprule
\makecell{Spectral\\Supervision} & \makecell{Causal\\Mask} & \makecell{Non-uniform\\Mapping} & FID$\downarrow$ & IS$\uparrow$ \\
\midrule
$\times$ & $\times$ & $\times$ & 6.30 & 190.1 \\
\checkmark & $\times$ & $\times$ & 5.64 & 255.1 \\
\checkmark & \checkmark & $\times$ & 3.49 & 222.6 \\
\checkmark & \checkmark & \checkmark & \textbf{3.02} & \textbf{282.2}\\
\bottomrule
\end{tabular}
}
\vspace{-5mm}
\end{table}

\subsection{Applications}
\label{subsec:app}
In this section, we provide a quantitative analysis for the potential applications of SpectralAR.
For super-resolution, we conduct our experiments based on the 256$\times$256 images generated by SpectralAR-$d16$ in Table~\ref{tab:main}, and directly upsample them to 512$\times$512 as the baseline.
We construct 4 sub-images using strided and patch-based methods for SpectralAR-Stride and -Patch, respectively.
We then upsample the 4 sub-images to 256$\times$256 and use SpectralAR to refine them in the spectral domain, and finally reassemble them to generate the final result.
According to Table~\ref{tab:app}, SpectralAR can indeed serve as a spectrum completer for the super-resolution task.
In addition, we also experiment with truncated autoregressive generation, where we discard the last few tokens.
For example, we discard the last 5 tokens in SpectralAR-Trunc.5 in Table~\ref{tab:app}.
The generation performance worsens slowly when the number of truncated tokens is fewer than 10, and therefore it is possible to further improve token efficiency through truncation according to the requirement for generation quality.

\subsection{Ablation Study}
We conduct ablation study to validate the effectiveness of our design choices in Table~\ref{tab:ablate}.
The first row corresponds to the baseline TiTok implementation for autoregressive generation.
Note that the spectral supervision alone could improve FID compared with the TiTok counterpart.
This is because the spectral design enhances sequence causality compared with the overall reconstruction target which ignores the correlation between tokens and thus complicates the autoregressive modeling process.
The spectral causal mask further enhances performance by improving causality in the encoding and decoding architecture.
And the non-uniform token-frequency mapping technique guides the model to focus more on the crucial low-frequency components while representing the high-frequency components efficiently, thus further improving performance.

\section{Conclusion}
\label{sec: conclusion}
In this paper, we have proposed the spectral autoregressive visual generation method for both causal and efficient autoregressive modeling of image data.
Specifically, we first convert images into 1D sequences with nested spectral tokenization, which supervises the 1D tokens with hierarchical spectral reconstruction targets to introduce causality into the sequence.
In addition, we have adopted causal masks for spectral tokens in the encoder and decoder networks to further enhance causality from the architectural perspective.
We have also designed a non-uniform token-frequency mapping with greater emphasis on the low-frequency components in order to improve token efficiency.
On the ImageNet-1K generation benchmark, our SpectralAR achieves superior performance in autoregressive visual generation compared with other methods.

\textbf{Limitations and future work.}
Although SpectralAR has achieved promising results, its scalability to larger models and datasets remains to be explored.
In addition, the applications of SpectralAR in videos generation, image restoration and forgery detection are also interesting topics.

{
    \small
    \bibliographystyle{ieeenat_fullname}
    \bibliography{main.bbl}
}

\end{document}